# Automatic Detection of Small Groups of Persons, Influential Members, Relations and Hierarchy in Written Conversations Using Fuzzy Logic

French Pope III [1], Rouzbeh A. Shirvani [1], Mugizi Robert Rwebangira[2], Mohamed Chouikha[1], Ayo Taylor[2], Andres Alarcon Ramirez[1], Amirsina Torfi[1], french.pope@bison.howard.edu, Rouzbeh.asghari@gmail.com, rweba@scs.howard.edu, mchouikha@howard.edu, ayo.taylor@gmail.com, alarcon27@hotmail.com, amirsina.torfi@ bison.howard.edu

[1]Electrical and Computer Engineering, Howard University, Washington, D.C. 20059 USA
[2]Systems and Computer Science, Howard University, Washington, D.C. 20059 USA

*Abstract—* Nowadays a lot of data is collected in online forums. One of the key tasks is to determine the social structure of these online groups, for example the identification of subgroups within a larger group. We will approach the grouping of individual as a classification problem. The classifier will be based on fuzzy logic. The input to the classifier will be linguistic features and degree of relationships (among individuals). The output of the classifiers are the groupings of individuals. We also incorporate a method that ranks the members of the detected subgroup to identify the hierarchies in each subgroup. Data from the HBO television show *The Wire* is used to analyze the efficacy and usefulness of fuzzy logic based methods as alternative methods to classical statistical methods usually used for these problems. The proposed methodology could detect automatically the most influential members of each organization *The Wire* with 90% accuracy.

## I. INTRODUCTION

In the last decade there has been increasing use of online platforms such as opinion forums, chat groups, and social networks because of broad access to the internet and people's communication needs. This new way of communicating has allowed people with different customs, cultures, and locations to get together virtually to interact and sometimes cooperate around common interests. On the other hand, the motivation of many e-commerce companies for understanding the behavior of internet users as well as the interest of some security agencies for detecting security threats has created the need for analyzing the data generated by online communities. In addition, because of the massive use of online communication tools and large amount of information generated by their users, it is almost impossible to manually analyze all of the generated information. Therefore, there have lately been important efforts that seek to automatically analyze and extract relevant information from written data corresponding to dialogues among several persons. One of the active areas of research is to detect associations among the members of an online community by subgroup identification in written conversations. The idea of subgroup identification is to identify members from a community who have similar ways of thinking or have the same affiliation and may cooperate each other. Yessenalina et al. [1] proposed a methodology that classifies the speaker's side in a corpus of congressional floor debates, using the speaker's final vote on the bill as a labeling for side. This work infers agreement between speakers based on cases where one speaker mentions another by name, and a simple algorithm for determining the polarity of the sentence in which the mention occurs. Gupte et al. [2] address the problem of segmenting small group meetings in order to detect different group configurations in an intelligent environment. They propose an unsupervised method based on the calculation of the Jeffrey divergence between histograms of speech activity observations. These histograms are generated from adjacent windows of variable size slid from the beginning to the end of a meeting recording. Elson et al. [3] proposed a method for detecting social networks from nineteenth-century British novels and serials. They linked two characters based on whether or not they conversed.

Tan et al. [4] proposed an algorithm that seeks to detect groups of people in Twitter with the same affiliation. To do this, it assumes that connected users are more likely to hold similar opinions. Finally, the discussants were classified in groups based on how often they reply to each other. Kunegis et al. [5] studied user relationships in the Slashdot technology news site. Slashdot gives users the option of tagging other users as friends or foes, providing positive and negative endorsements. Abu et al. [6] identified subgroups in ideological discussions. To do this, they identified the discussion participants, comments, and the reply structure of the thread (i.e. who replies to whom). Then, they used sentiment analysis to determine the polarity of the comment (positive or negative) made by a particular participant. Finally, to identify the subgroup membership of each discussant, they use the fact that the attitude profiles of discussants who share the same opinion are more likely to be similar to each other than to the attitude profiles of discussants with opposing opinions. Hassan et al. [7] take into consideration the posts exchanged between participants and sentiment analysis to build a signed network representation of the discussion. After building the signed network representation of discussions, the large group of discussants split into many subgroups with coherent opinions.

Also of great interest is the identification of the hierarchy of the members from a particular subgroup. The hierarchy of a group is important because it allows us to detect the most influential members from a group as well as the role and importance of each member in a group. In the case of identifying influential members, Rienks [8] proposed an algorithm for detecting influencers in a corpus of conversations. He focuses entirely on non-linguistic behavior and looks at (verbal) interruptions and topic initiations. Brdiczka et al. [9] proposed a method for deciding for each participant in a thread whether or not he or she is an influencer in that particular thread, this approach relies on identification of three types of conversational behavior: persuasion, agreement/disagreement, and dialog patterns. In the same way, Clauset et al. [10] used Markov Chain Monte Carlo sampling to estimate the hierarchical structure in a network. Gupte et al. [2] proposed a measure of hierarchy in a directed online social network, and proposed an algorithm to compute this measure.

The fuzzy logic approach to determining groupings of individuals in written conversation extracts features from conversations and determines through fuzzy logic the likelihood that individuals are in the same group. The approach then considers those who communicate with each other coupled with the previous results to increase the accuracy of the grouping. The fuzzy logic method allows for weight and values to be assigned to features displayed through written speech as well as taking into account who is in a conversation. The approach relies on empirical data that is extracted from the written conversations to determine the grouping of each individual. The counting of features provides a way to move from qualitative space to quantitative space which enables the measurement of the distance between characters and assigning them to different groups. Features are used as input into the fuzzy logic algorithm which groups individuals in conversation based on the empirically used features.

   This paper is organized as follows. Section 2 describes the proposed methodology to extract influential members from a diverse group of persons. Additionally, it also describes a method to identify small subgroups of people as well as close relations of the members of small subgroups. In the same way, Section 3 describes how to find the relationship matrix between characters. Section 4 tries to describe the data used in the experiments and the results obtained. Finally, Section 5 shows the conclusions and future works.

## II. Feature Extraction

Extracting features in this context amounts to identifying linguistic characteristics of individuals under consideration. This is performed in two steps, using the LightSide tool [12] for identifying the common characteristics of a group using feature vectors, and then reducing the feature vectors to a minimum of independent characteristics. LightSide is an open source text mining and machine learning tool that can extract frequency of word usage and parts of speech features to predict membership in certain groups. LightSide is used in this case to extract frequently displayed features by multiple individuals within the text. "The Wire" text was used and the goal was to classify individuals by their place in the hierarchy, whether they are Gang, Police, or Informant. By informant, we mean that person is connected to both gang and police, but he/she is a gang member or used to be a gang member. In order to do that, we should have some initial knowledge about each group so that we can extend our knowledge with a learning algorithm. So we first, marked four characters in each group that we already know that whether they are police or gang members so we could have some initial knowledge. This is where LightSide comes in, this software is able to extract the features of each labeled character. It goes through all of the characters that are labeled (as police or gang members) and extracts the part of speech (POS) features that we are going to deal with for them. As a definition, each part of speech refers to a category to which a word is assigned in accordance with its syntactic function. In English the main parts of speech are noun, pronoun, adjective, determiner, verb, adverb, preposition, conjunction, and interjection. For instance, the part of speech distribution in the sentence below is as follows:

This (determiner) student (noun) is (auxiliary verb) working (verb) on (preposition) an (determiner) interesting (adjective) project (noun).

The way the features are dealt with is general and they can be extended to more general cases, because the method is not based on some specific words and based on the structure of the text. Now two categories of feature are gathered, the first one is the initial police POS features ($F_{1P}$) and the second one is the initial gang POS features ($F_{1G}$). Both groups can have some features in common, but some filters are applied to and reduce the features to the ones that are specific to each group making the features independent. Independence by the two groups not having features in common and the two feature spaces not having features in common. It is shown in equation (1) that $F_{1P}$ is the initial gang POS features, and $F_{1G}$ is the initial gang POS features, after reduction the feature space will change to $F_P$ and $F_G$. The two feature spaces, $F_{1P}$ and $F_{1G}$, may have some features in common, but $F_P$ and $F_G$ are fully independent feature vector space and orthogonal to each other.

$$\exists\ f_{1P}\in F_{1P}: f_{1P}\in F_{1G} \quad \text{or} \quad F_{1P}\cap F_{1G}\neq\emptyset \qquad (1)$$

$$\exists\ f_{1G}\in F_{1G}: f_{1G}\in F_{1P} \quad \text{or} \quad F_{1P}\cap F_{1G}\neq\emptyset \qquad (2)$$

$$\text{If } f_G\in F_G\rightarrow f_G\notin F_P \quad \text{or} \quad F_P\cap F_G=\emptyset \qquad (3)$$

$$\text{If } f_P\in F_P\rightarrow f_P\notin F_G \quad \text{or} \quad F_P\cap F_G=\emptyset \qquad (4)$$

Now, the police POS features and gang POS features are characterized by separate groups. It is the time to go through the other characters that have no information about them. In other words, extend the algorithm throughout the whole text and get some information about the unknown characters. The unknown characters are the ones with undetermined group affiliation. The labels are removed from the characters that were labeled previously and the text is run through LightSide. LightSide will extract all possible POS feature for each characters. Each character might have thousands of POS features, but among them only the ones common with $F_P$ and $F_G$ are needed. As a result, the feature space is reduced.

At the next step, we assigned two values for each character, one for the number of times they show $F_P$ and the other for the number of times they show $F_G$. Each person is assigned two values of A and B in which,

A=Number of times character shows $F_G$
B=Number of times character shows $F_P$

Equation (5) and (6) show the ratio that each character shows $F_G$ and $F_P$.

$$a=\frac{A}{A+B} \quad (5)$$

$$b=\frac{B}{A+B} \quad (6)$$

### III. RELATIONSHIP DETERMINATION

The objective of our approach is to explore the relationships of individuals in conversation and use this as another factors in finding characters of the same group. The approach used to examine relationships is developed out of the need to determine who is communicating with whom and assign values to those in conversations together. Values are assigned to each individual in a conversation by taking the previous individuals in the conversation as well as those who follow. There are two methods to determine the values which are to assign a true or false value to those in conversation or to simply count the number of times that there is a back and forth in the conversation. A vector was created for each individual in the conversations. The vector consists of N columns for each of the N individuals in the total text. The approached takes the individual currently talking and assigns a 1 to the individuals directly before and after. This would mean that the characters are in direct conversation according to the text. The algorithm also assigns the value 0.5 to person that proceeds and follows the individual by two for indirect contact. Suppose the part of the conversation is in the order of Table I. Consider Person 1, as connected to person 2 directly one time and indirectly one time so a value of 1.5 is assigned for the relation between Person 1 and Person 2. The same can be done for the relation between Person 1 and 3, they are connected three times directly which means they are repeated right after each other three times and they are connected indirectly one time which will assign the value of 0.5 to their connection. The total value of relation between Person 1 and Person 3 would be 3.5. Without seeing the individuals who are in conversation this algorithm yields insight as to those who are related simply by when they speak. The next step is to aggregate the results of the line by line conversation and create a vector for each person that represents the relationships developed throughout the text.

Fuzzy logic is the tool used to deal with the crisp numbers of equation (5) and (6), by that the numbers are made into some interpretable values.

TABLE I.    PART OF A CONVERSATION

| Conversation |
|---|
| Person 1 |
| Person 2 |
| Person 3 |
| Person 1 |
| Person3 |
| Person 1 |
| Person 4 |
| Person 1 |

In Table II the relation matrix related to Table I is shown, the higher the value in the matrix is considered as a higher weight in the relationship between the two individuals. Relation matrix is named R and is a N×N matrix which is a symmetric matrix.

TABLE II.    RELATION MATRIX OF TABLE 1

| Convers. | Person 1 | Person 2 | Person 3 | Person 4 |
|---|---|---|---|---|
| Person 1 |  | 1+0.5=1.5 | 0.5+1+1+1= 3.5 | 1+1=1 |
| Person 2 | 1+.5=1.5 |  | 1 |  |
| Person 3 | 0.5+1+1+1= 3.5 | 1 |  | .5 |
| Person 4 | 1+1=2 |  | .5 |  |

### IV. FUZZIFICATION

Fuzzy logic is the tool used to deal with the crisp numbers of equation (5) and (6), by that the numbers are made into some interpretable values. The fuzzy logic algorithm used is based on c-mean fuzzy logic clustering algorithm [13]. With three different groups (police, gang, and informant), fuzzy logic approach helps to assign membership values for each person to each group. In this paper we are trying to use the fuzzy C-Means algorithm, but it has a little difference with C-Means method. The difference is that it is not iterative and the center of each cluster (group) is known, so there is no need to iterate. The C-Means algorithm is a method to calculate the degree of membership for each person to each group. The basic and challenging part in fuzzy logic is the rules. This algorithm helps us to reduce the rules in a great deal which leads to a faster process. We have three major rules in our two fuzzy logic boxes that are as follows;

1. If a=1 and b=0 and d=A-B=15 character is gang.
2. If a=0 and b=1 and d=A-B=-15 character is police.
3. If a=0.5 and b=0.5 and d=A-B=0 character is informant.

Different characters have different a, b, and d value, we compare their closeness to each group based on these three values. If their values are closer to each of the above rules, they will have a higher membership value to that group. According to equation (7) algorithm developed allows to calculate the degree of membership of each character to each group (See Figure 1).

$$\mu_{G_i}(x) = \frac{1}{\sum_{j=1}^{3} \frac{\|x-F_i\|^2}{\|x-F_j\|^2}} \quad 1 \leq i \leq 3, \ x \in X \quad (7)$$

$G_i$ is one of the three groups, in this study $G_1$ is considered for gang, $G_2$ for police, and $G_3$ for the informant case. Vector $F_i$ is the center of each group in which the values are assigned based on empirical observation from dataset, knowing that there are three groups implies that there will be three centers of the groups $F_1$, $F_2$, and $F_3$. In other words $F_1$, $F_2$, and $F_3$ show ideal values for a person to be gang, police, and informant respectively. X is the dataset, and $x$ is one member in the dataset which is the normalized feature vector that yields some values for each character in order to compare them together and see how close they are to each other or how far they are from the center of each group ($F_1$, $F_2$, $F_3$). Note that each character has a normalized vector x. As a result, we are able to calculate how close each character is to the ideal gang, police, and informant. There will be three values ($\mu_{G_1}$, $\mu_{G_2}$, $\mu_{G_3}$) for each person that gives information about the membership of each person to each of the groups. Indeed, these three numbers are stored in each row of $M_1$ for each character. Each row shows the membership of each character in gang, police, and informant group. The same approach can be done for the second fuzzy logic box and the membership values are named $\mu_{G_{i2}}$ (stored in $M_2$) in which i=1, 2, 3 refers to gang, police, and informant respectively. This algorithm reduces the overlearning process and CPU time since this clustering (grouping) method will reduce our rules drastically.

## V. RESULTS

Up to now, the algorithm that is used in this paper has been described, now to implement the algorithm on the dataset. The overall flow diagram of the method is shown in Figure 1. Parts that are inside the dashed line is our black box that processes the input data and gives the output as the degree of membership of being gang, police, and informant for each character. As it was said in previous section, for each character there will be three values for $\mu_{G_{i1}}$ and three values for $\mu_{G_{i2}}$ in which;

$$\mu_{G_{11}} + \mu_{G_{21}} + \mu_{G_{31}} = 1 \ , \quad \mu_{G_{12}} + \mu_{G_{22}} + \mu_{G_{32}} = 1 \quad (8)$$

The $\mu_{G_{i1}}$ values for all characters are collected in one matrix and called $M_1$, also the same for $\mu_{G_{i2}}$ and call it $M_2$. By considering the total number of characters as N, $M_1$ and $M_2$ are shown in equation (9).

$$M_1 = \begin{bmatrix} \mu_{G_{11}}^1 & \mu_{G_{21}}^1 & \mu_{G_{31}}^1 \\ \mu_{G_{11}}^2 & \mu_{G_{21}}^2 & \mu_{G_{31}}^2 \\ \vdots & \vdots & \vdots \\ \mu_{G_{11}}^N & \mu_{G_{21}}^N & \mu_{G_{31}}^N \end{bmatrix} \quad M_2 = \begin{bmatrix} \mu_{G_{12}}^1 & \mu_{G_{22}}^1 & \mu_{G_{32}}^1 \\ \mu_{G_{12}}^2 & \mu_{G_{22}}^2 & \mu_{G_{32}}^2 \\ \vdots & \vdots & \vdots \\ \mu_{G_{12}}^N & \mu_{G_{22}}^N & \mu_{G_{32}}^N \end{bmatrix} \quad (9)$$

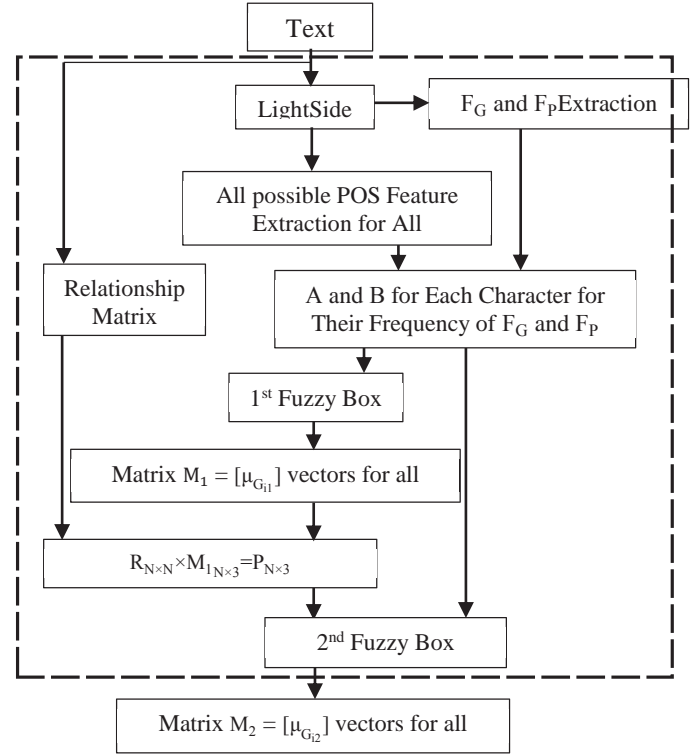

Figure 1. Flow diagram of the method

The characters that are going to be dealt with are the main characters in "The Wire". The output result after applying the first and second fuzzy box to the characters and their text are according to the Table IV, V. As it can be seen in the, the first fuzzy box was not successful to extract the exact rule of some characters. It was after using relation matrix and second fuzzy box that we were able to extract the rule of each character as a gang, police, and informant with high accuracy. Table III compares our output results with the k-mean statistical based method. It can be seen that k-mean statistical based method is able to do so with 85 percent accuracy, but we were able to extract the rule of all characters with high accuracy. The initial selection of the most influential people among a group of persons based on the number of comments made by the participants in a conversation yielded the results shown in the Table VI.

TABLE III. COMPARISON BETWEEN THE RESULT OF STATISTICAL BASED METHOD AND FUZZY LOGIC BASED METHOD

| Character | K-Mean Statistical Based Method | | Fuzzy Logic Based Method | |
|---|---|---|---|---|
| | Accuracy | Overall accuracy | Accuracy | Overall accuracy |
| Avon | ✓ | | ✓ | |
| Stringer | ✓ | | ✓ | |
| McNulty | ✓ | | ✓ | |
| Carver | ✓ | | ✓ | |
| Freamon | ✓ | | ✓ | |
| Greggs | × | | ✓ | |
| Dee | ✓ | 85% | ✓ | 100% |
| Omar | ✓ | | ✓ | |
| Bunk | ✓ | | ✓ | |
| Daniels | ✓ | | ✓ | |
| Russel | ✓ | | ✓ | |
| Nick | ✓ | | ✓ | |
| Sobotka | × | | ✓ | |
| Ziggy | ✓ | | ✓ | |

TABLE IV. THE RESULT FIRST FUZZY LOGIC BOX

| Characters | Results After First Fuzzy Box | | | |
|---|---|---|---|---|
| | $\mu_{G11}$ | $\mu_{G21}$ | $\mu_{G31}$ | Accuracy |
| Bey | 0.14 | 0.03 | 0.81 | ✓ |
| Avon | 1 | 0 | 0 | ✓ |
| Stringer | 1 | 0 | 0 | ✓ |
| Phelan | 0.00 | 0.99 | 0.00 | ✓ |
| McNulty | 0 | 1 | 0 | ✓ |
| Pearlman | 0.00 | 0.99 | 0.00 | ✓ |
| Carver | 0.00 | 0.99 | 0.00 | ✓ |
| Freamon | 0.00 | 0.99 | 0.00 | ✓ |
| Greggs | 0.00 | 0.99 | 0.00 | ✓ |
| Dee | 0.17 | 0.17 | 0.65 | ✓ |
| Omar | 0.00 | 0.76 | 0.22 | × |
| Bunk | 0 | 1 | 0 | ✓ |
| Norris | 0.00 | 0.01 | 0.98 | × |
| Daniels | 0 | 1 | 0 | ✓ |
| Landsman | 0.00 | 0.99 | 0.00 | ✓ |
| Prez | 0.00 | 0.99 | 0.00 | ✓ |
| Burrel | 0.00 | 0.99 | 0.00 | ✓ |
| Russel | 0 | 1 | 0 | ✓ |
| Nick | 1 | 0 | 0 | ✓ |
| Sobotka | 1 | 0 | 0 | ✓ |
| Ziggy | 0.99 | 0.00 | 0.00 | ✓ |

Table VI shows the members who make the highest number of comments in a conversation where 251 persons participates. The Table VI also shows the affiliation of each selected person, and it is marked with red color the persons involved in criminal activities, whereas it is highlighted in green the members with no criminal activity. The accumulated value of all comments made by the members shown in the Table VI accounts for the 90% of the total number of comments in the 10 episodes of the TV show, The Wire. Finally, the methodology based on the number of comments recovers the most important characters of the TV show. The first group shown in the Table VI is constituted mainly by police officers or persons who enforce the law. The only character classified in this group who is not a police officer is Mr. Sobotka. During the TV show, he makes arrangements with European gangsters to smuggle illegal goods through the Baltimore's port. On the other hand, the second group is constituted mostly by persons involved with criminal activities, that is, drug dealers, smugglers, etc. However, a total of 4 characters who are police officers were misclassified in the second group primarily related with criminal activities. In order to identify the hierarchy or the grade of importance of the members that constitute a group of persons, it is analyzed three distinct groups of people present in the TV show, The Wire.

TABLE V. THE RESULT SECOND FUZZY LOGIC BOX

| Characters | Results After First Fuzzy Box | | | |
|---|---|---|---|---|
| | $\mu_{G11}$ | $\mu_{G21}$ | $\mu_{G31}$ | Accuracy |
| Bey | 0.81 | 0.07 | 0.10 | ✓ |
| Avon | 0.98 | 0.00 | 0.00 | ✓ |
| Stringer | 0.96 | 0.01 | 0.01 | ✓ |
| Phelan | 0.02 | 0.82 | 0.15 | ✓ |
| McNulty | 0.01 | 0.96 | 0.02 | ✓ |
| Pearlman | 0.00 | 0.96 | 0.02 | ✓ |
| Carver | 0.02 | 0.82 | 0.15 | ✓ |
| Freamon | 0.01 | 0.94 | 0.04 | ✓ |
| Greggs | 0.02 | 0.83 | 0.14 | ✓ |
| Dee | 0.00 | 0.00 | 0.99 | ✓ |
| Omar | 0.03 | 0.33 | 0.63 | ✓ |
| Bunk | 0.00 | 0.96 | 0.02 | ✓ |
| Norris | 0.00 | 0.98 | 0.01 | ✓ |
| Daniels | 0.00 | 0.96 | 0.02 | ✓ |
| Landsman | 0.01 | 0.94 | 0.03 | ✓ |
| Prez | 0.01 | 0.90 | 0.07 | ✓ |
| Burrel | 0.01 | 0.90 | 0.07 | ✓ |
| Russel | 0.00 | 0.98 | 0.00 | ✓ |
| Nick | 0.97 | 0.01 | 0.01 | ✓ |
| Sobotka | 0.89 | 0.04 | 0.06 | ✓ |
| Ziggy | 0.96 | 0.01 | 0.01 | ✓ |

TABLE VI.    IDENTIFICATION OF THE MOST INFLUENTIAL MEMBERS AMONG A GROUP OF PERSONS

| Members | Affiliation | # of comments |
|---|---|---|
| McNulty | Police Officer | 373 |
| BUNK | Police Officer | 238 |
| NICK | Involved in Criminals Activities | 228 |
| SOBOTKA | Involved in Criminals Activities | 191 |
| FREAMON | Police Officer | 175 |
| STRINGER | Top Drug Dealer | 164 |
| DANIELS | Police Officer | 148 |
| ZIGGY | Involved in Criminals Activities | 146 |
| AVON | Top Drug Dealer | 126 |
| RUSSELL | Port Authority Officer | 120 |
| GREGGS | Police Officer | 114 |
| DEE | Drug Dealer | 92 |
| OMAR | Involved in Criminals Activities | 82 |
| CARVER | Police Officer | 77 |
| VALCHEK | Police Commander | 70 |
| HERC | Police Officer | 56 |
| PREZ | Police Officer | 53 |
| SPIROS | Involved in Criminals Activities | 52 |
| LANDSMAN | Police Officer | 50 |
| ELENA | Police Officer | 50 |
| BODIE | Drug Dealer | 49 |
| LEVY | Attorney | 47 |
| PEARLMAN | leading Assistant State's Attorney | 44 |
| RAWLS | Police Officer | 41 |
| HORSEFACE | Involved in Criminals Activities | 40 |

That is, the police officers, and two organizations related to criminal activities. One of the criminal groups is the Barksdale organization which is led by Avon Barksdale and Stringer Bell. This criminal organization is responsible for multiples crimes and is the most powerful and violent crew in the Baltimore area. The other criminal group is the Sobotka family that is a Polish American Baltimore family. The head of the family is Frank Sobotka, a treasurer for the local union at the Baltimore docks. However, He is also involved along with his family in arrangements with criminals to smuggle illegal goods through the port. Thus, the Sobotka family not only has extensive connections to the Baltimore port, but also links to the criminal underworld. The hierarchies of the main members that constitute the three existing groups are shown in Table VII. The Table VII shows the main members of the distinct organizations present in TV show, The Wire. Additionally, the heads of each organization are highlighted in green color; in the same way, the mid-level and low-level members are colored in yellow and red respectively. On the other hand, the proposed methodology also seeks to automatically detect the hierarchy of the members that constitute a particular group of persons.

To do that, it takes into account the following features: average value of coordination, number of formulated questions, use of modal verbs, number of hedge, use of profanity, and number of terms of address. In order to compare the rank made by the proposed methodology with actual hierarchy of the members that belong to a particular group, it is used the squared difference of the obtained ranking and the actual ranking such as follows.

$$e = \frac{3}{n^3 - n} \sum_{k=1}^{n} (R_k - E_k)^2 \quad , \quad n > 1 \quad (10)$$

Where, n, is the number of members of the group being analyzed, the parameters $R_k$ and $E_k$ are the actual ranking and the obtained ranking respectively. The parameter e, called ranking error, takes values that range from 0 to 1, where a value of 0 means no error in the ranking, and a value of 1 is the maximum error.

TABLE VII.    HIERARCHIES OF THE THREE EXISTING GROUPS

| Police Officers | Barksdale organization (Criminals) | Sobotka family (Docks) |
|---|---|---|
| 1.Daniels (Deputy Commissioner) 2.Freamon (Detective) 3.McNulty (Detective) 4.Bunk (Detective) 5.Greggs (Detective) 6.Carver (Detective) 7.Russell (Port Authority Police Officer) | 1.Avon (Kingpin) 2.Stringer (Kingpin) 3.Bey (Soldier) 4.D'angelo (Dealer) 5.Bodie (Dealer) 6.Poot (Dealer) | 1.Sobotka (Head of the family) 2.Nick (Sobotka's Nephew) 3.Ziggy (Sobotka's Son) 1.Sobotka (Head of the family) 2.Nick (Sobotka's Nephew) 3.Ziggy (Sobotka's Son) |

TABLE VIII.    OVERALL OBTAINED RANKING OF THE SEASON 2 OF THE WIRE

| Police | Criminals | Docks |
|---|---|---|
| Freamon | Avon | Ziggy |
| Daniels | Stringer | Sobotka |
| McNulty | Bodie | Nick |
| Bunk | D'angelo | |
| Russell | Bey | |
| Carver | Poot | |
| Greggs | | |

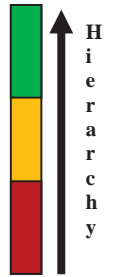

Hierarchy

Table VIII shows the overall ranking of the members in the three existing groups after averaging the positions of each member.

VI. CONCLUSION

We have explored the possibility of using fuzzy logic in computational linguistics to determine characteristics from text. For the particular case that we studied we found that indeed fuzzy logic can be a powerful method with high accuracy that out performs other methods in clustering and subgroup identification. One important aspect for future work is more extensive testing on different corpuses of data.